\documentclass[letterpaper, 10 pt, conference]{ieeeconf}  
\usepackage{amsmath,graphicx}
\usepackage{times}
\usepackage{epsfig}
\usepackage{graphicx}
\usepackage{amssymb}
\usepackage{url}
\usepackage{adjustbox}
\usepackage{booktabs}
\usepackage{caption}
\usepackage{comment}
\usepackage{graphicx}
\usepackage{mathtools}
\usepackage{multirow}
\usepackage{multicol}
\usepackage{xspace}
\usepackage{xfrac}
\usepackage{pifont}
\usepackage{gensymb}
\usepackage{algorithm}
\usepackage{algorithmic}
\usepackage{afterpage}

\usepackage{adjustbox}
\usepackage{amsmath}
\usepackage{amssymb}
\usepackage{booktabs}
\usepackage[style=base]{caption}
\usepackage{graphicx}
\usepackage{multirow}
\usepackage{multicol}
\usepackage{xspace}
\usepackage[table]{xcolor}

\makeatletter
\let\NAT@parse\undefined
\DeclareRobustCommand\onedot{\futurelet\@let@token\@onedot}
\def\@onedot{\ifx\@let@token.\else.\null\fi\xspace}

\makeatother

\usepackage[colorlinks,pagebackref=false,citecolor=blue,bookmarks=false,hypertexnames=true]{hyperref}

\mathchardef\mhyphen="2D 

\def\##1{\relax\ifmmode\mathchoice
{\mbox{\boldmath$\displaystyle#1$}}
{\mbox{\boldmath$\textstyle#1$}}
{\mbox{\boldmath$\scriptstyle#1$}}
{\mbox{\boldmath$\scriptscriptstyle#1$}}\else
\hbox{\boldmath$\textstyle$}\fi}

\title{
Ensemble-based Semi-supervised Learning to Improve \\ Noisy Soiling Annotations in Autonomous Driving}

\author{Michal U\v{r}i\v{c}\'{a}\v{r}$^{1}$, 
Ganesh Sistu$^{2}$, Lucie Yahiaoui$^{2}$ and Senthil Yogamani$^{2}$
\\
{\normalsize $^{1}$Independent Researcher, Czech Republic \quad
$^{2}$Valeo Vision Systems, Ireland}
}

\begin{document}
%
\maketitle
%


\begin{abstract}
Manual annotation of soiling on surround view cameras is a very challenging and expensive task. The unclear boundary for various soiling categories like water drops or mud particles usually results in a large variance in the annotation quality. As a result, the models trained on such poorly annotated data are far from being optimal. In this paper, we focus on handling such noisy annotations via pseudo-label driven ensemble model which allow us to quickly spot problematic annotations and in most cases also sufficiently fixing them. We train a soiling segmentation model on both noisy and refined labels and demonstrate significant improvements using the refined annotations. It also illustrates that it is possible to effectively refine lower cost coarse annotations. 
\end{abstract}



\section{Introduction}

\begin{figure}[tb]
\centering
\includegraphics[width=0.48\textwidth]{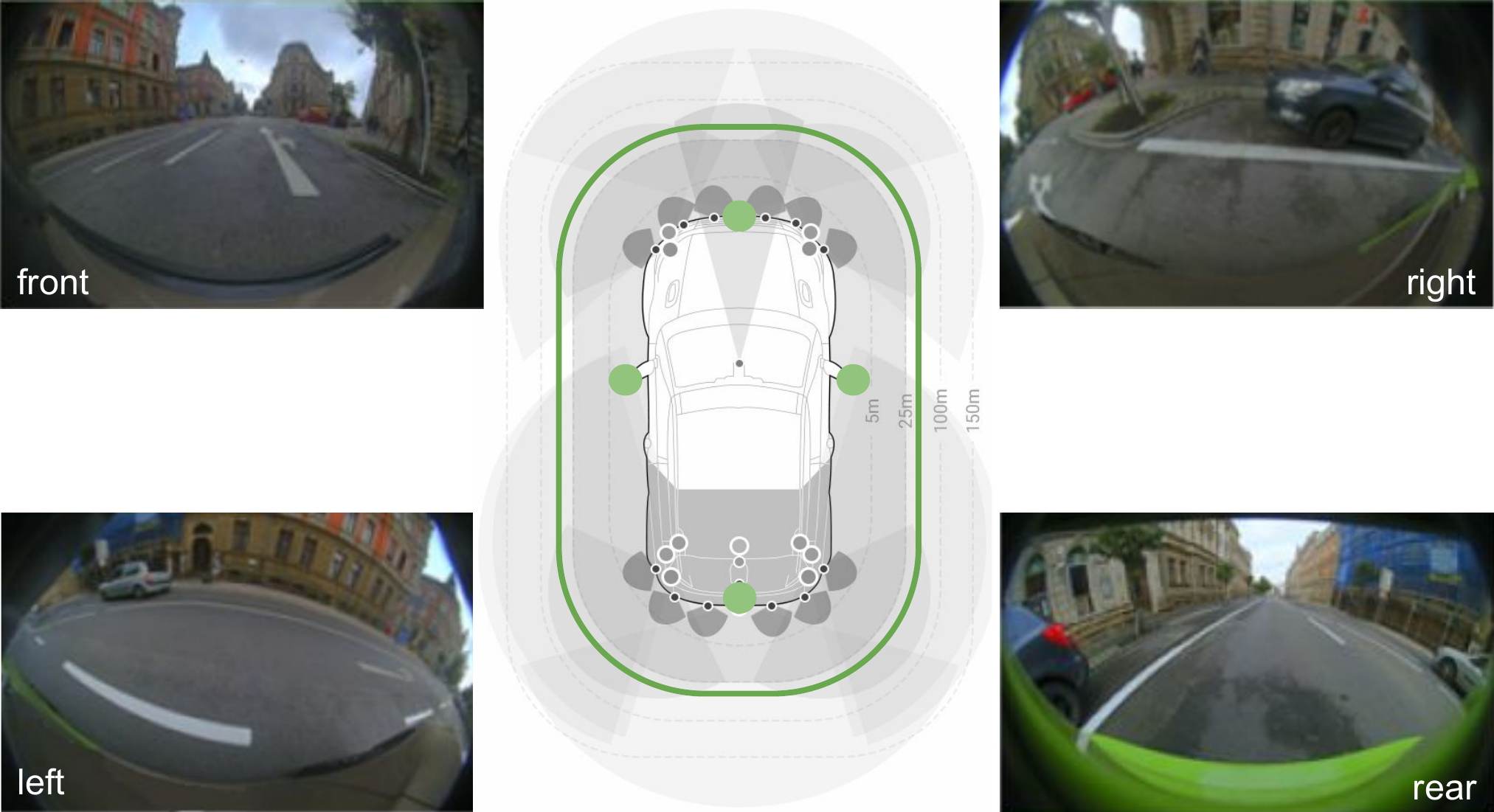}

\vspace{1.5mm}

\includegraphics[width=0.48\textwidth]{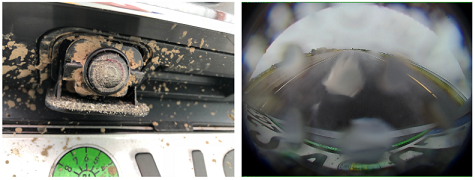}
\caption{Top: Surround view cameras and their mounting positions on the vehicle. Bottom: Soiled camera lens due to mud and a water soiled image.}
\label{fig:svs}
\end{figure}

Wide-angle cameras are important sensors for certain tasks in autonomous driving, such as automated parking. Figure~\ref{fig:svs} shows how these cameras are organized in order to get a 360\degree{} view around the vehicle. Since these cameras are usually placed externally and exposed to the environmental conditions, one needs to account for the reliability of these sensor input in order to achieve a full autonomy. It is not uncommon that during harsh weather (such as heavy rain, snow, etc.) or when driving in off-road conditions, the surround view cameras get dirty (Figure~\ref{fig:svs} bottom). 
In general, there is limited work on fisheye camera perception for tasks such as object detection \cite{rashed2021generalized}, depth estimation \cite{kumar2018monocular}, SLAM \cite{tripathi2020trained} and multi-task learning \cite{kumar2021omnidet}.
Recently, the problem of dealing with camera soiling in autonomous driving has gained increased attention~\cite{uricar2019soilingnet, Uricar-2019a, uricar2019desoiling, maddu2019fisheyemultinet}.

Automated driving systems perform well in normal weather conditions and one of its main challenges is to work effectively in adverse weather conditions. Particularly, visual perception algorithms degrade severely in adverse weather conditions. Adverse conditions can cause soiling of camera lens which causes even more severe degradation in performance. Relatively, there is little work addressing to handle this scenario. Porav et al.~\cite{porav2019i} implemented a de-raining algorithm using PatchGAN-based architecture and a clever data collection mechanism to collect rainy and non-rainy images of the same scene. Sakaridis et al.~\cite{sakaridis2018semantic} uses synthetically generated foggy scenes to evaluate the impact of degradation of object detection and propose to defog in order to improve performance. Ki et al~\cite{Ki-2018} propose to use conditional GAN to perform dehazing.

While some work pointed out the problems with manual annotation process and offered artificial generation of training data with pixel-precise annotation~\cite{Uricar_2021_WACV, 10.1007/978-3-030-58565-5_27}, there is not much work dedicated to solving the issue of fixing existing imprecise manual annotations or some method to quickly spot the images with annotation errors in large scale datasets.
The problem with imprecise annotations is two-fold. Firstly, the classification models learned with the full supervision inherently reflect the quality of the annotations provided in the training. Secondly, the evaluation of the trained models is done on imprecise annotations, so the actual performance analysis is clouded and it can easily happen that a best performing model on \emph{real} data is discarded due to poorer performance on the imprecisely annotated test data. 

\begin{figure}[t]
    \centering
    \includegraphics[width=0.5\textwidth]{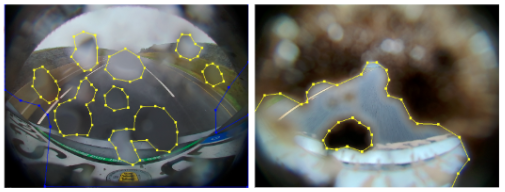}
    \caption{Soiling annotation using polygons. Boundaries are ambiguous and difficult to mark. }
    \label{fig:soilingannotation}
\end{figure}

The soiling classification and localization task is problematic due to the unclear boundary of individual soiling categories and vague definition of soiling classes, such as when the soiling is already considered to be in the transparent or opaque category. Moreover, due to the annotation acquisition costs, coarse polygonal contours are commonly used instead of precise segmentation masks.  Figure~\ref{fig:soilingannotation} illustrates the difficulties of doing manual annotation and shows how ambiguous and subjective it can be.

Ideally, the solution would be to design a model which accounts for the known problems of annotation precision and operates in a semi-supervised manner. However, despite the enormous progress in semi-supervised learning~\cite{van2020survey}, this is still considered as the machine learning holy grail and far from being applied reliably in the real world scenarios, such as the autonomous driving. The other way of approaching this problem is to design a semi-automatic way of improving the annotation quality and reduce its inconsistencies. 

We selected the latter approach and propose to deal with the problem via the concept of pseudo-label~\cite{lee2013pseudo} ensemble~\cite{zheng2019new}. Pseudo-labeling is an interesting concept of handling imprecise annotations (e.g. prediction of some classifier) for semi-supervised learning. The pseudo-label for a data sample is usually chosen based on the maximum predicted probability and then considered to be the real label in the standard training loop. This is equivalent to entropy regularization~\cite{lee2013pseudo}. Because of the nature of manual annotations which we use, we consider all labels to be pseudo-labels and train the ensemble classifier in two consecutive stages. First stage selects the pseudo-label randomly, whereas the second stage picks it using the nearest neighbor principle (to the output of the first stage classification). The output from the second stage classifier provides refined annotations, which can be readily used for the standard supervised learning.







\section{Proposed Method} \label{sec:proposed}

The soiling task was first formally defined in~\cite{uricar2019soilingnet} based on WoodScape dataset~\cite{Yogamani_2019_ICCV}. WoodScape provides segmentation annotations $\mathcal{T} = \left\{ \#x_i, \#y_i \right\}_{i=1}^{n}$. Where $\#x \in \mathcal{I}^{H \times W}$ denotes the image of a fixed resolution 
and 
$\#y \in \mathcal{Y}$ denotes the per-pixel annotation of 
a set of four classes, namely $\mathcal{C} \in \{ \mathrm{opaque}, \mathrm{semi\mhyphen transparent}, \mathrm{transparent}, \mathrm{clean}\}$. 
The class \emph{opaque} means that in that particular region, it is impossible to see through, \emph{semi-transparent} and \emph{transparent} are blurry regions where background colors are visible with diminished texture. The difference between the  \emph{semi-transparent} and \emph{transparent} classes is the ability to recognize the object through the blur. The last class is \emph{clean} which indicates no soiling present. We follow the same labelling categories and definitions in our work.

Inspired by~\cite{zheng2019new}, we propose a two stage approach for learning an ensemble classifier using the pseudo-label (PL) concept introduced by~\cite{lee2013pseudo}. The ensemble classifier is a neural network, consisting of two separate encoders and one decoder. First encoder processes the input image, while the second encoder processes all channel-wise concatenated pseudo-labels to propagate the information from them. 

\begin{algorithm}[tb]
\centering
\caption{Two-stage ensemble learning} \label{alg:algo}
\begin{algorithmic}[1]
\REQUIRE $\#{x_i}, \mathrm{PL}_i = \{ f_1(\#x_i), f_2(\#x_i), \dots, f_9(\#x_i)  \},$ \\
$S(\mathrm{PL}_i), i = 1, \dots, n$, $m$ \\

\COMMENT{1st stage training} 
\WHILE{stopping criterion not met}
    \STATE \COMMENT{Random fit}
    \FOR {$i = 0$ \TO $m$}
        \STATE Sample $p \thicksim \mathcal{U}(1, n)$ \COMMENT{to select sample}
        \STATE Sample $q \thicksim \mathcal{U}(1, 9)$ \COMMENT{to select PL}
        \STATE Add sample $\{\#x_p, S(\mathrm{PL}_p), f_q(\#x_p) \} $ to mini-batch
    \ENDFOR
    \STATE Update 1st stage network $\mathcal{H}_1$ weights by back-propagating loss: $$\mathcal{L}(\#x, c) = -\log\left( \frac{\exp(x[c])}{\sum_j \exp(\#x[j])} \right)$$
    \STATE Empty mini-batch
\ENDWHILE

\COMMENT{2nd stage training}
\WHILE{stopping criterion not met}
    \STATE \COMMENT{Nearest neighbor fit}
    \FOR {$i = 0$ \TO $m$}
        \STATE Sample $p \thicksim \mathcal{U}(1, n)$ \COMMENT{to select sample}
        \STATE $\hat{y} = \mathcal{H}_1(\#x_p, S(\mathrm{PL}_p)$
        \STATE $\hat{q} = \arg\min_{q=1,\dots,9} \mathcal{L}\left(\hat{y}, f_q(\#x_p)\right)$
        \STATE Add sample $\{\#x_p, S(\mathrm{PL}_p), f_{\hat{q}}(\#x_p) \} $ to mini-batch
    \ENDFOR
    \STATE Update 2nd stage network $\mathcal{H}_2$ weights by back-propagating loss $\mathcal{L}(\#x, c)$
\ENDWHILE
\end{algorithmic}
\end{algorithm}

\begin{figure*}[t]
    \centering
    \includegraphics[width=\textwidth]{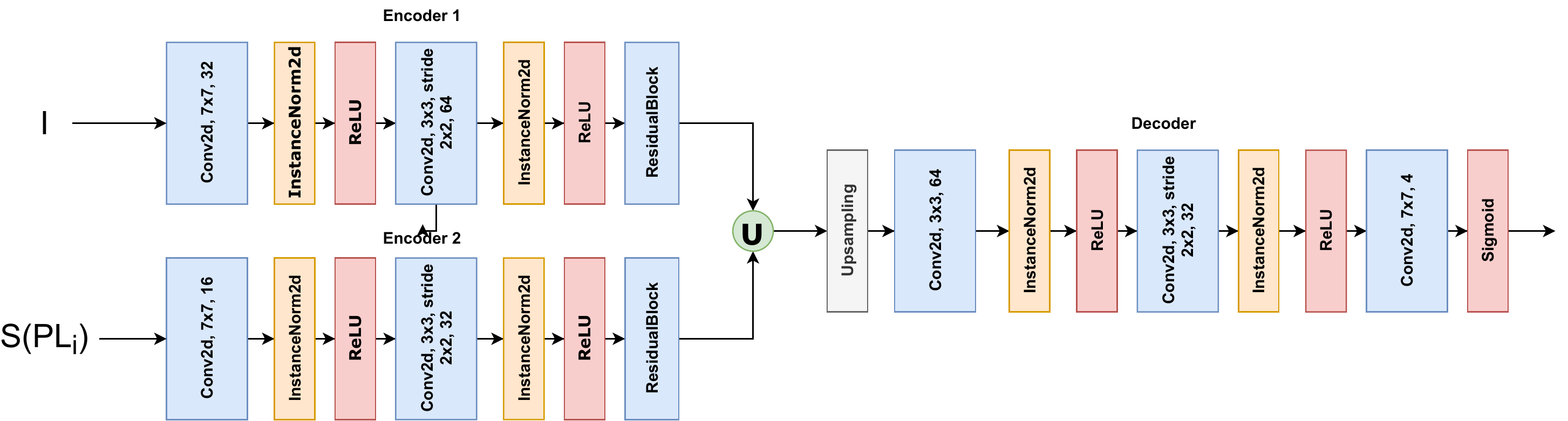}
    \caption{Proposed architecture. RGB image and channel-wise concatenated pseudo-labels (PL) are used as inputs. They are passed through two independent encoders. The output of both encoders is concatenated and passed to the decoder which produces refined soiling segmentation annotation.}
    \label{fig:EnsembleSAWSeg}
\end{figure*}

\begin{figure*}[tb]
\centering



\includegraphics[width=0.8\textwidth]{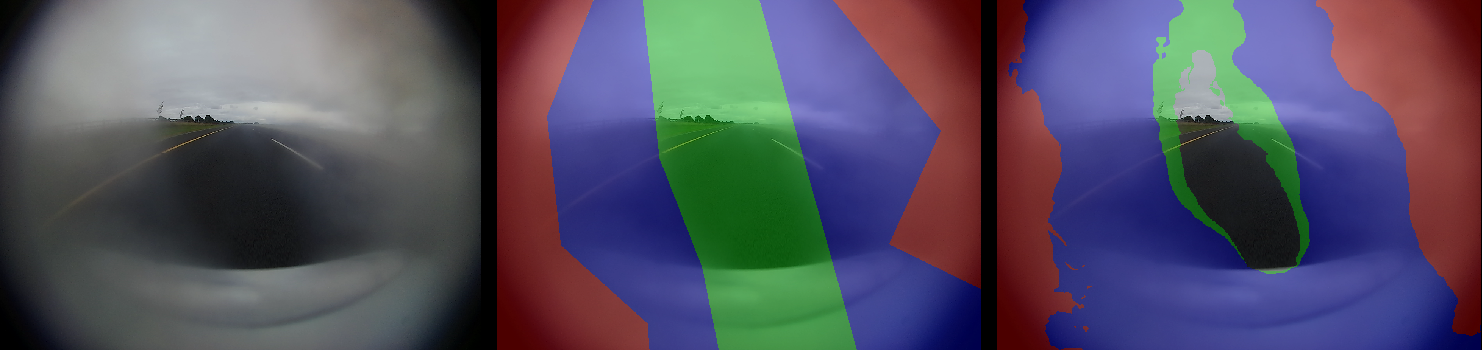}

\vspace{0.5mm}

\includegraphics[width=0.8\textwidth]{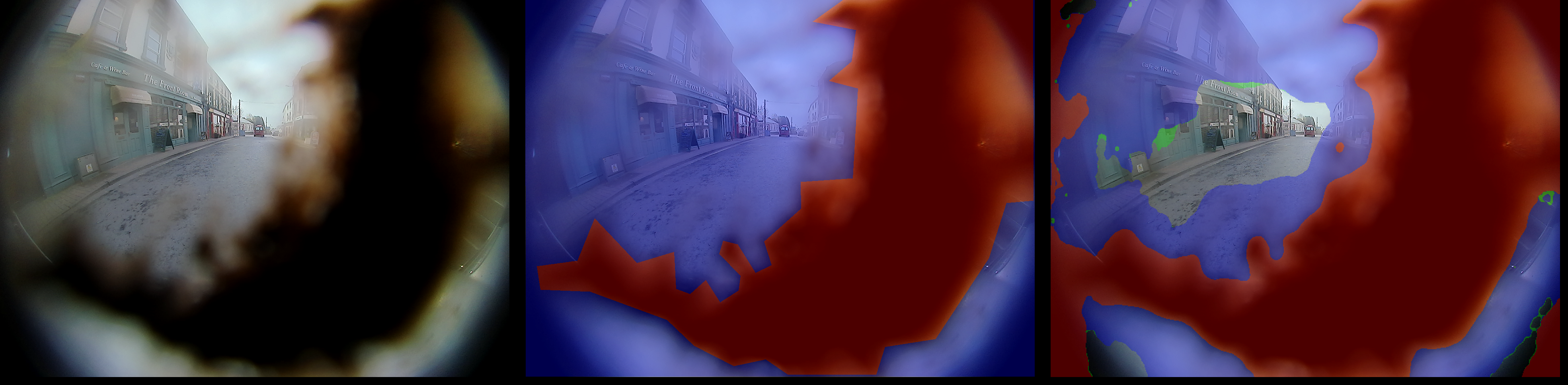}

\vspace{0.5mm}

\includegraphics[width=0.8\textwidth]{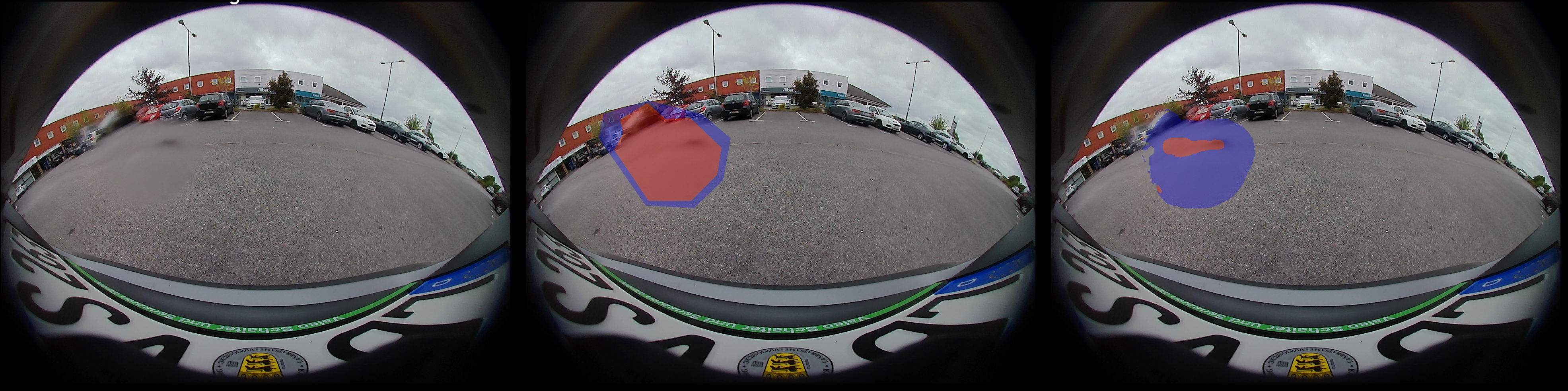}
\caption{Three sets of samples are presented. Left: Original image, Middle: Manual annotation and Right: Ensemble refined annotation. Class color coding: {\color{green}Green --- transparent}, {\color{blue}blue --- semi-transparent}, {\color{red} red --- opaque}, and original image color --- clean.  }
\label{fig:manual-vs-ensemble}
\end{figure*}

\begin{figure*}[tb]
\centering

\includegraphics[height=0.15\textheight]{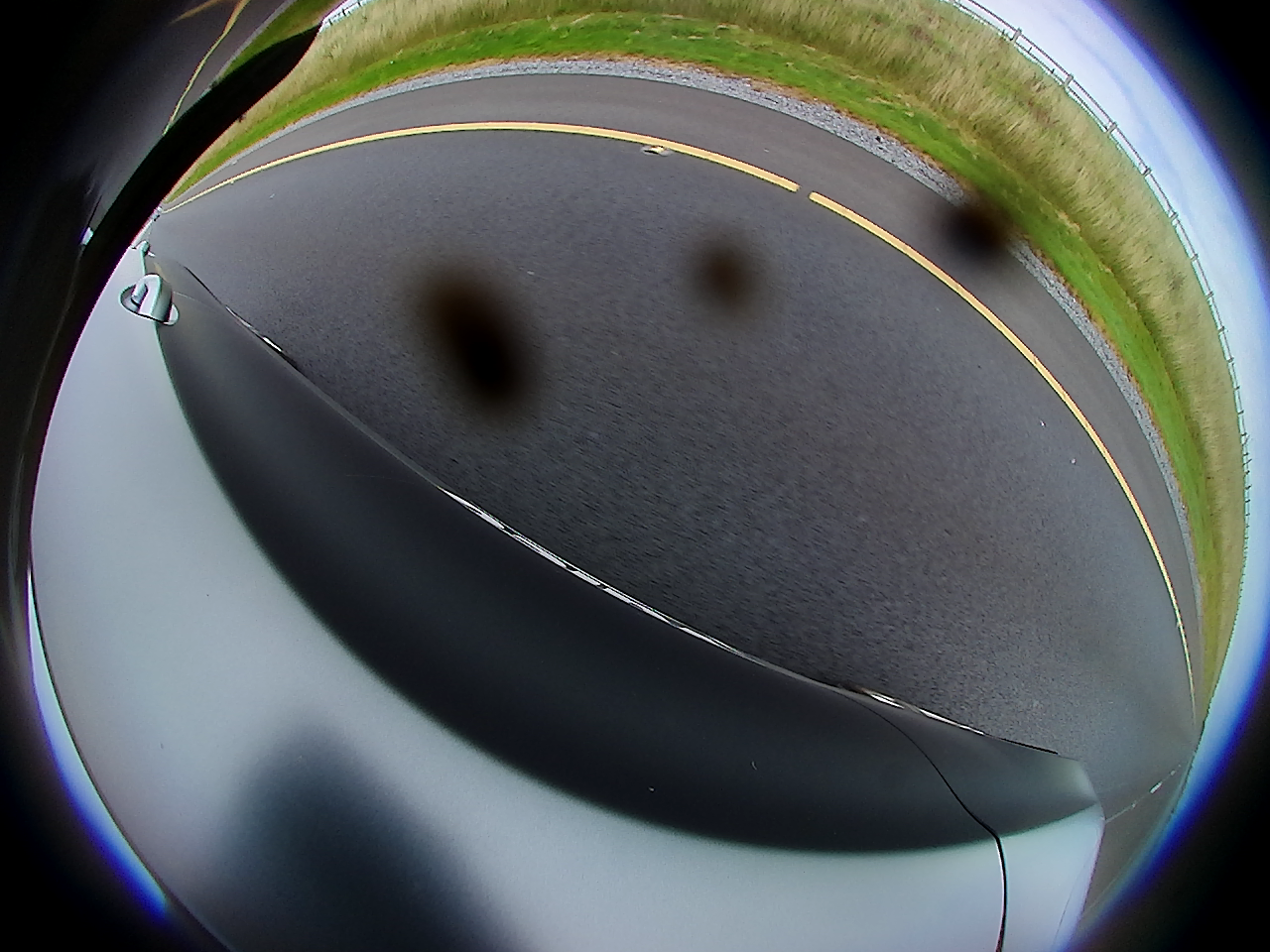}
\includegraphics[height=0.15\textheight]{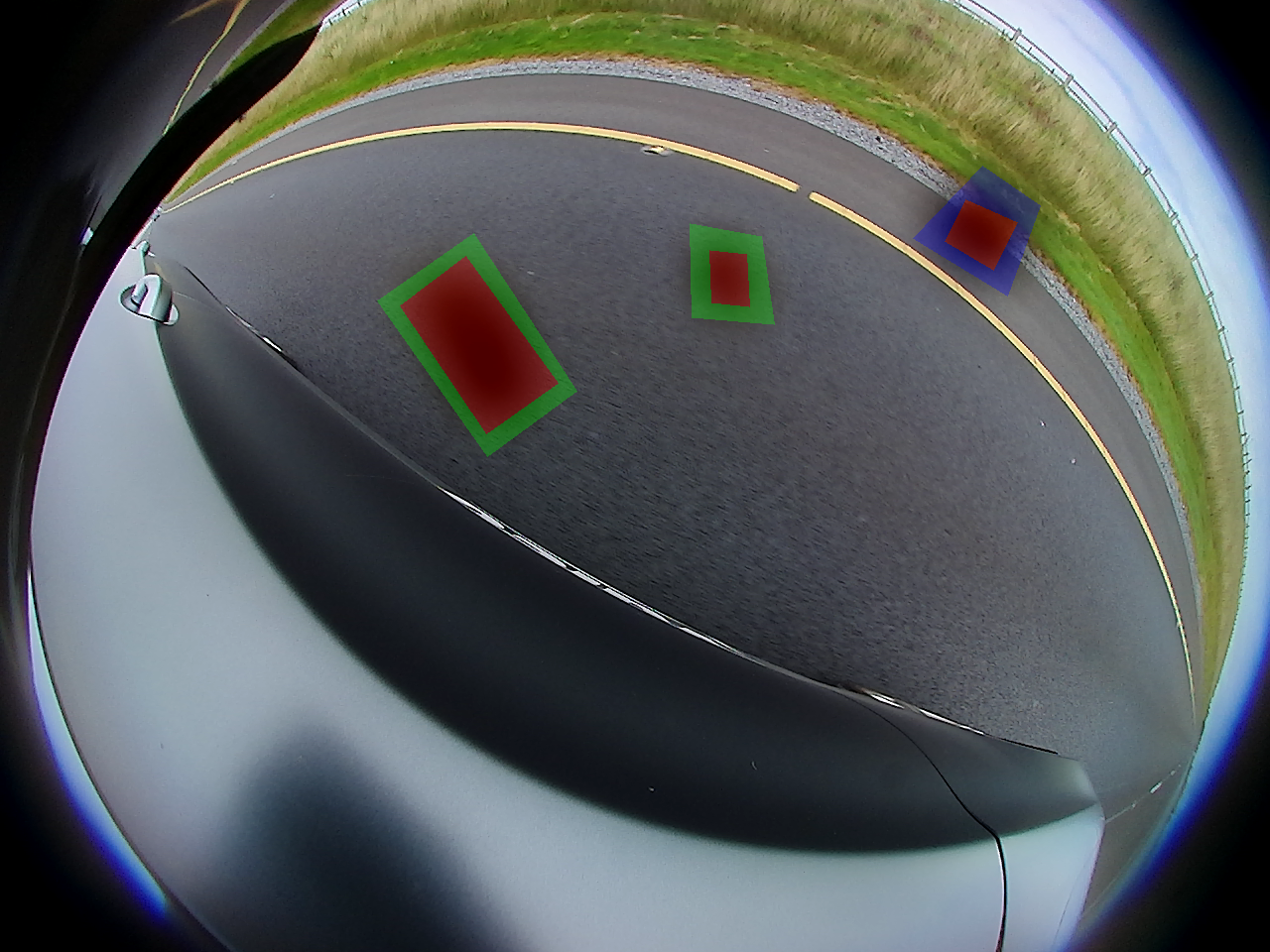}
\includegraphics[height=0.15\textheight]{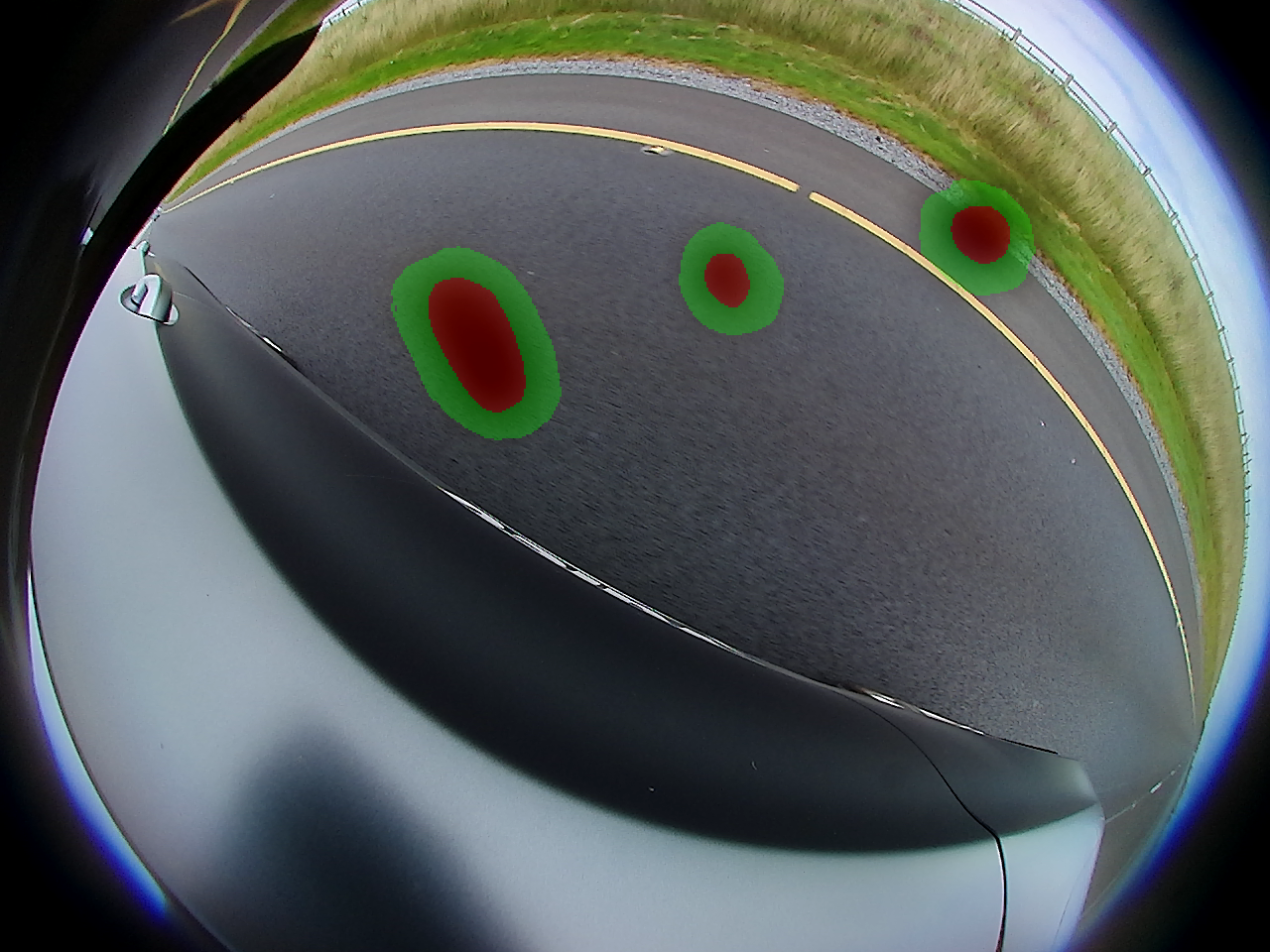}
\includegraphics[height=0.15\textheight]{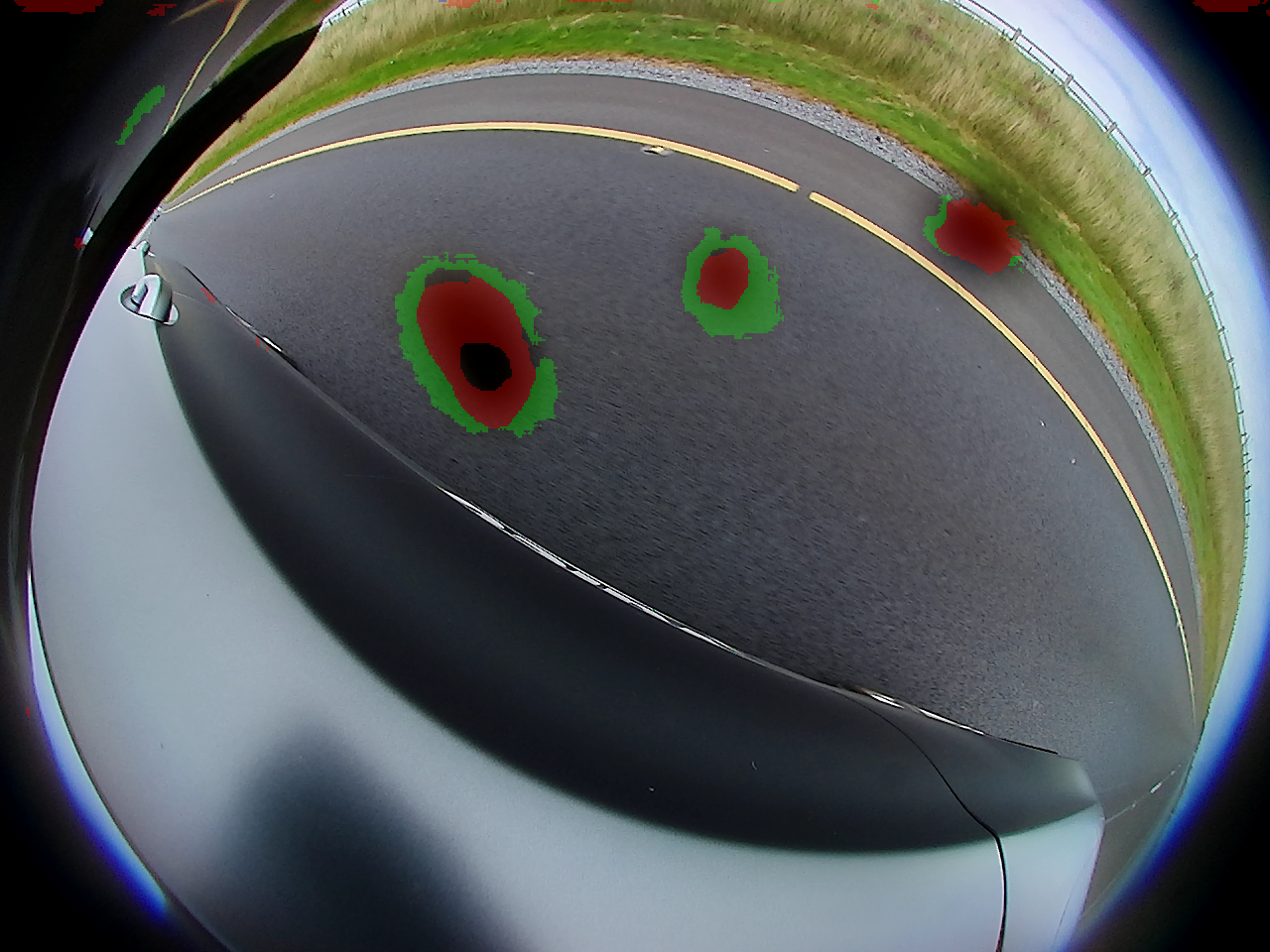}
\includegraphics[height=0.15\textheight]{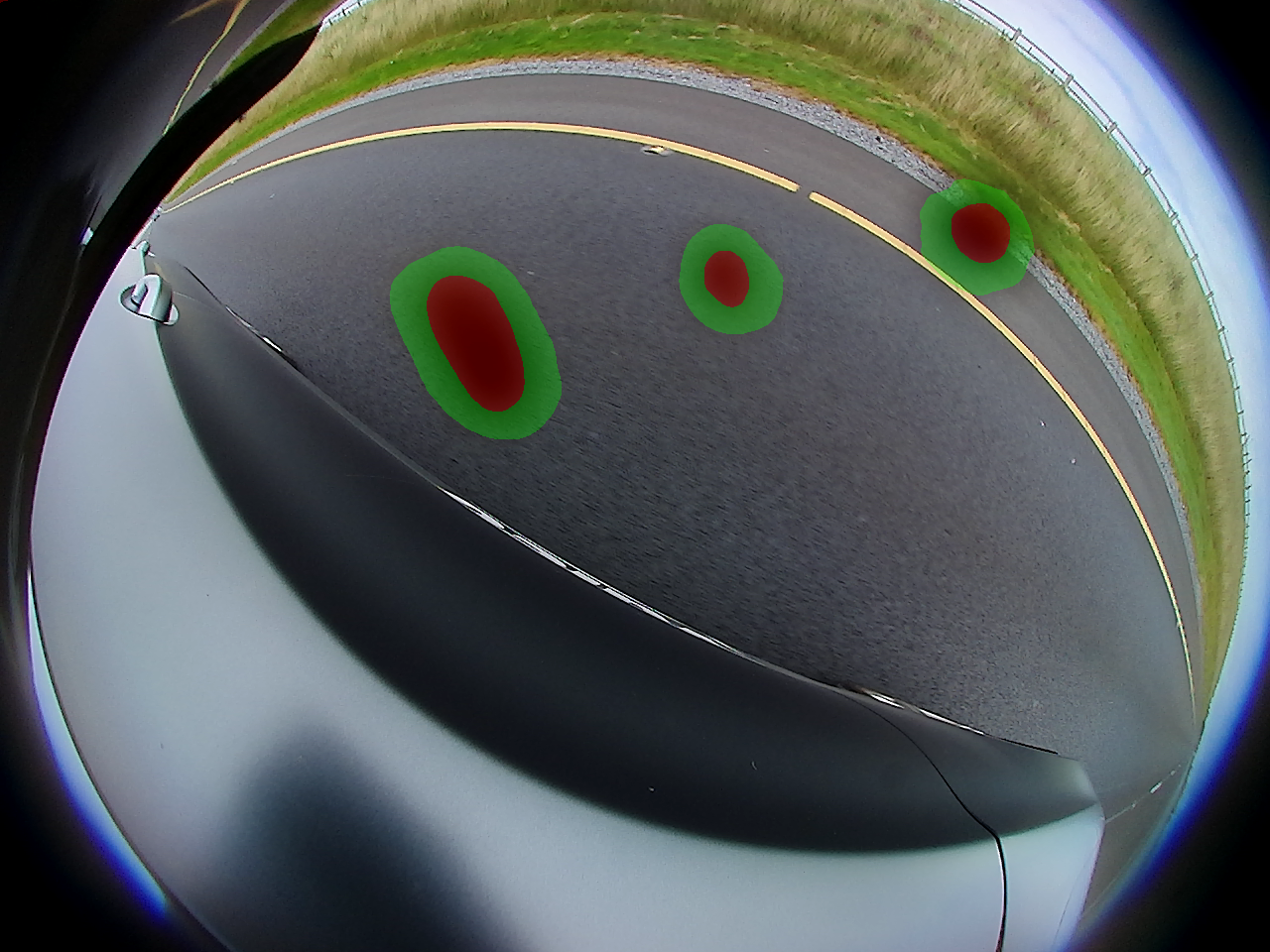}
\includegraphics[height=0.15\textheight]{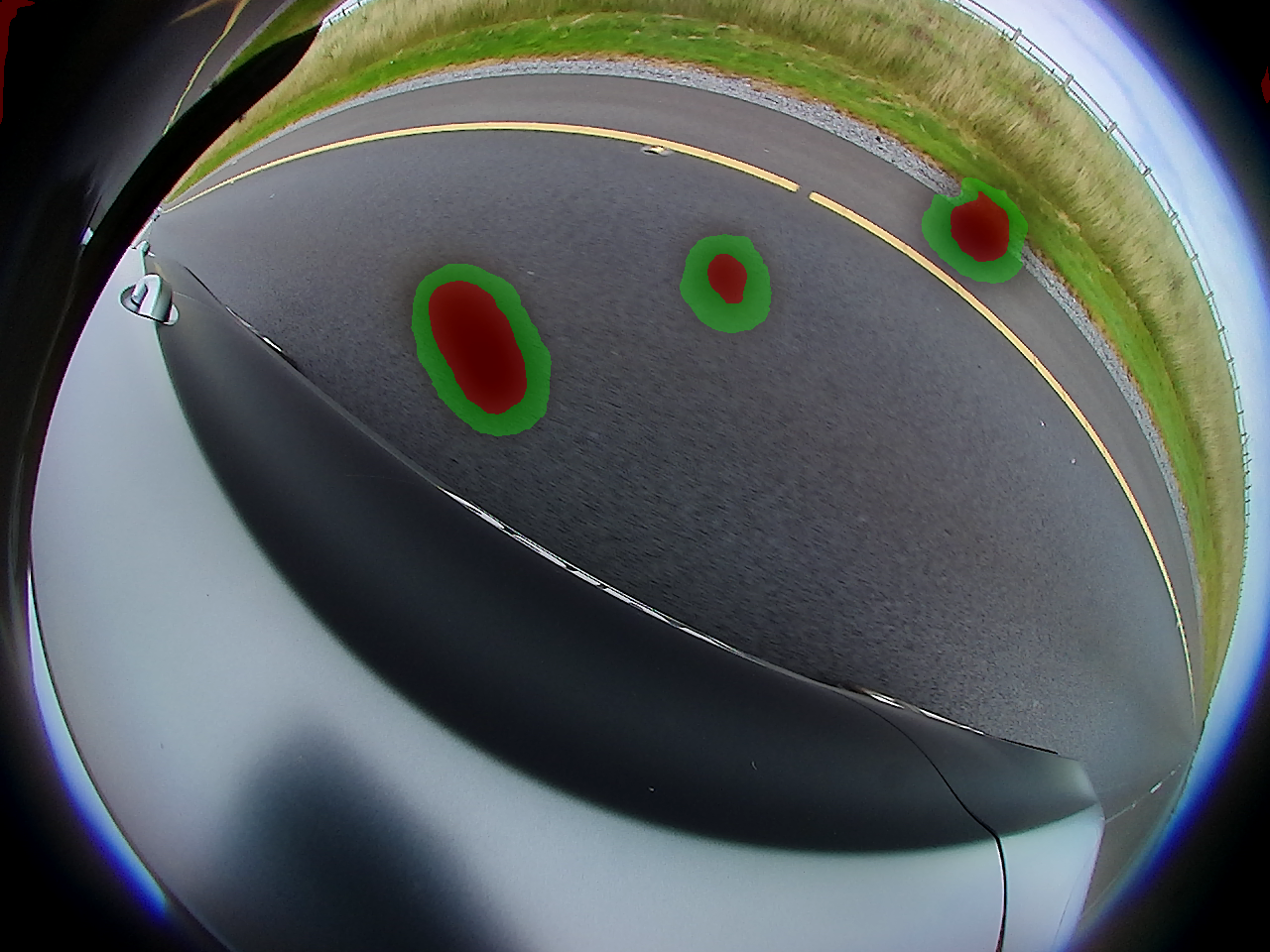}

\vspace{1.5mm}

\includegraphics[height=0.15\textheight]{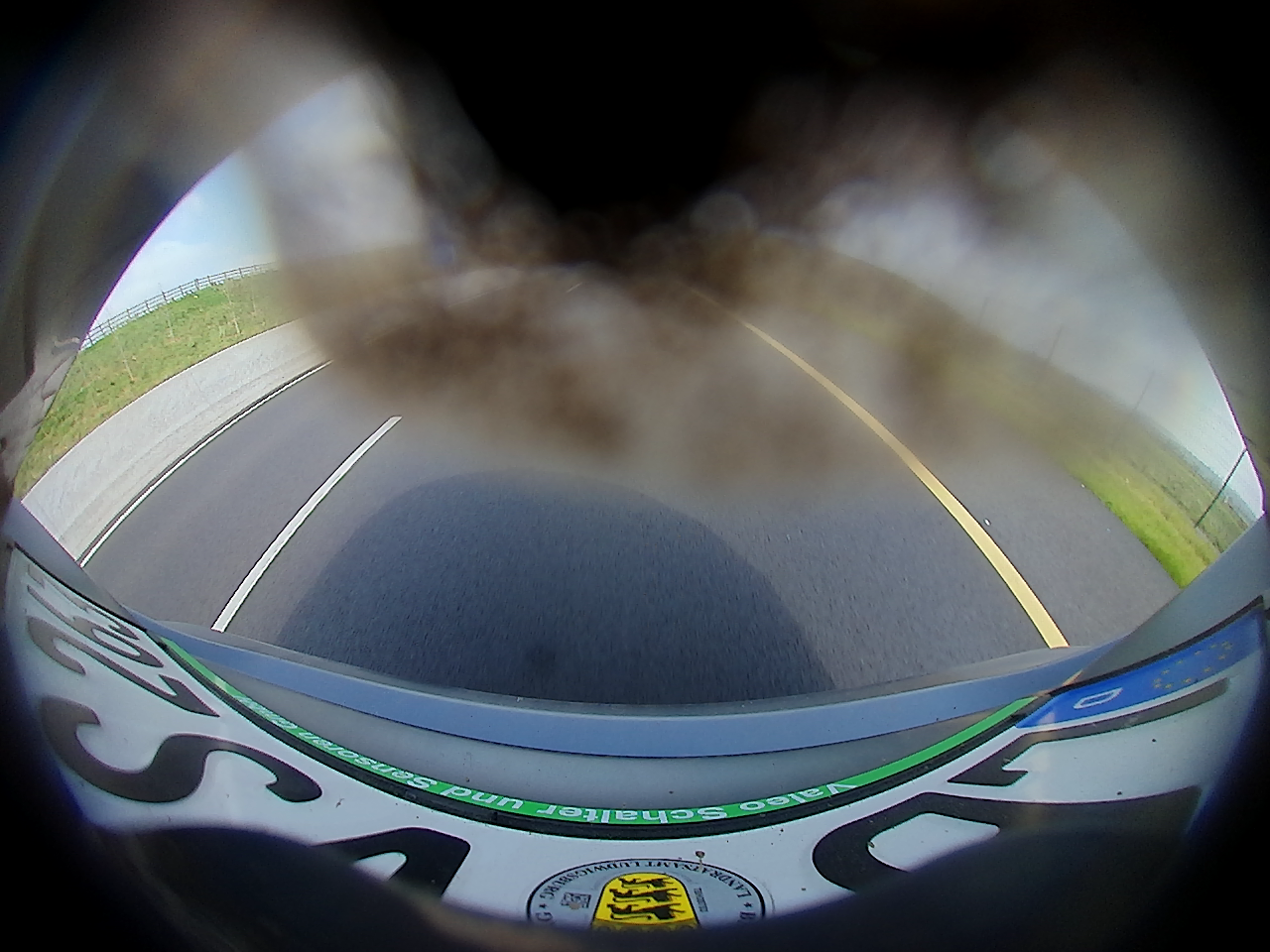}
\includegraphics[height=0.15\textheight]{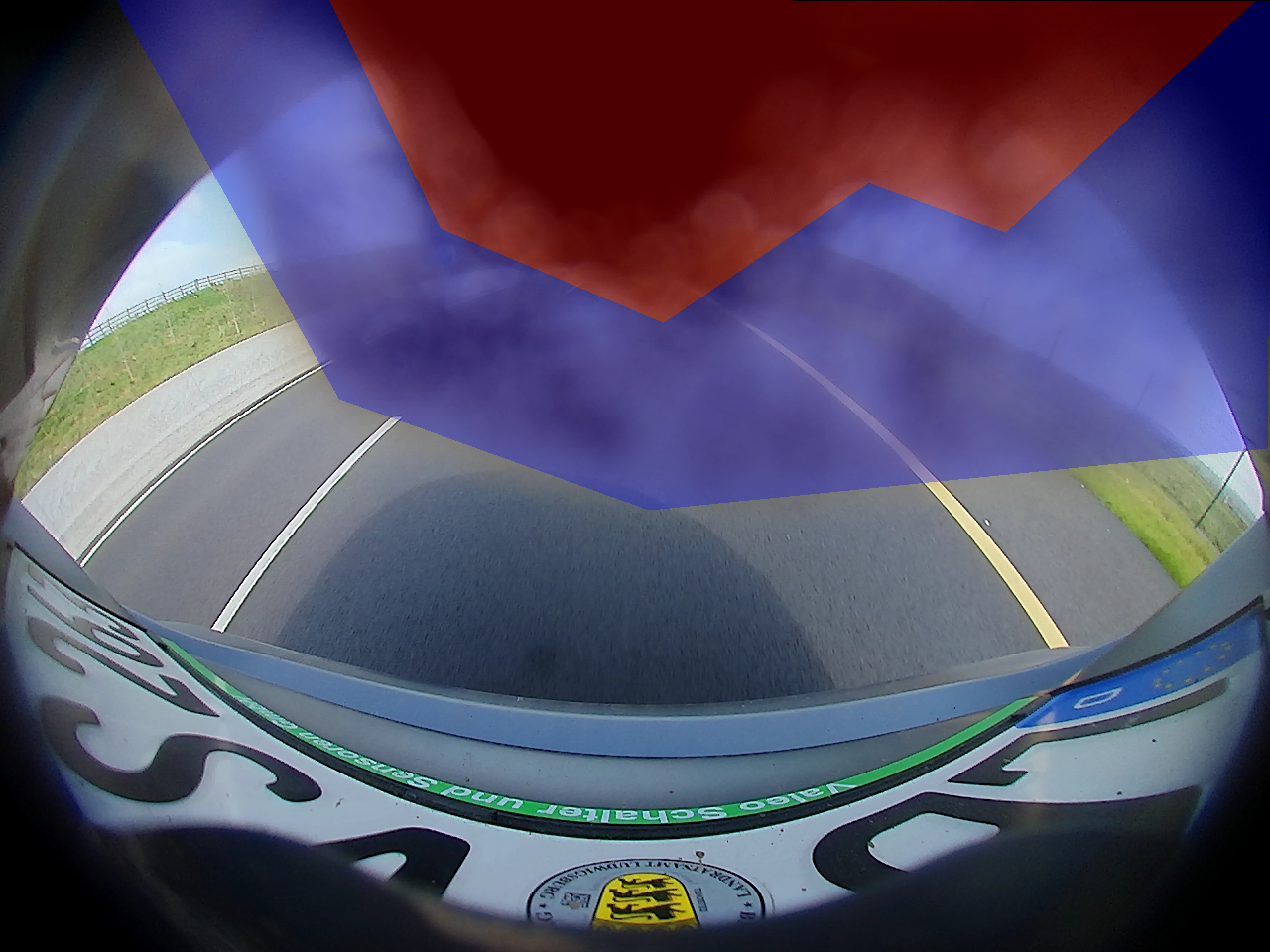}
\includegraphics[height=0.15\textheight]{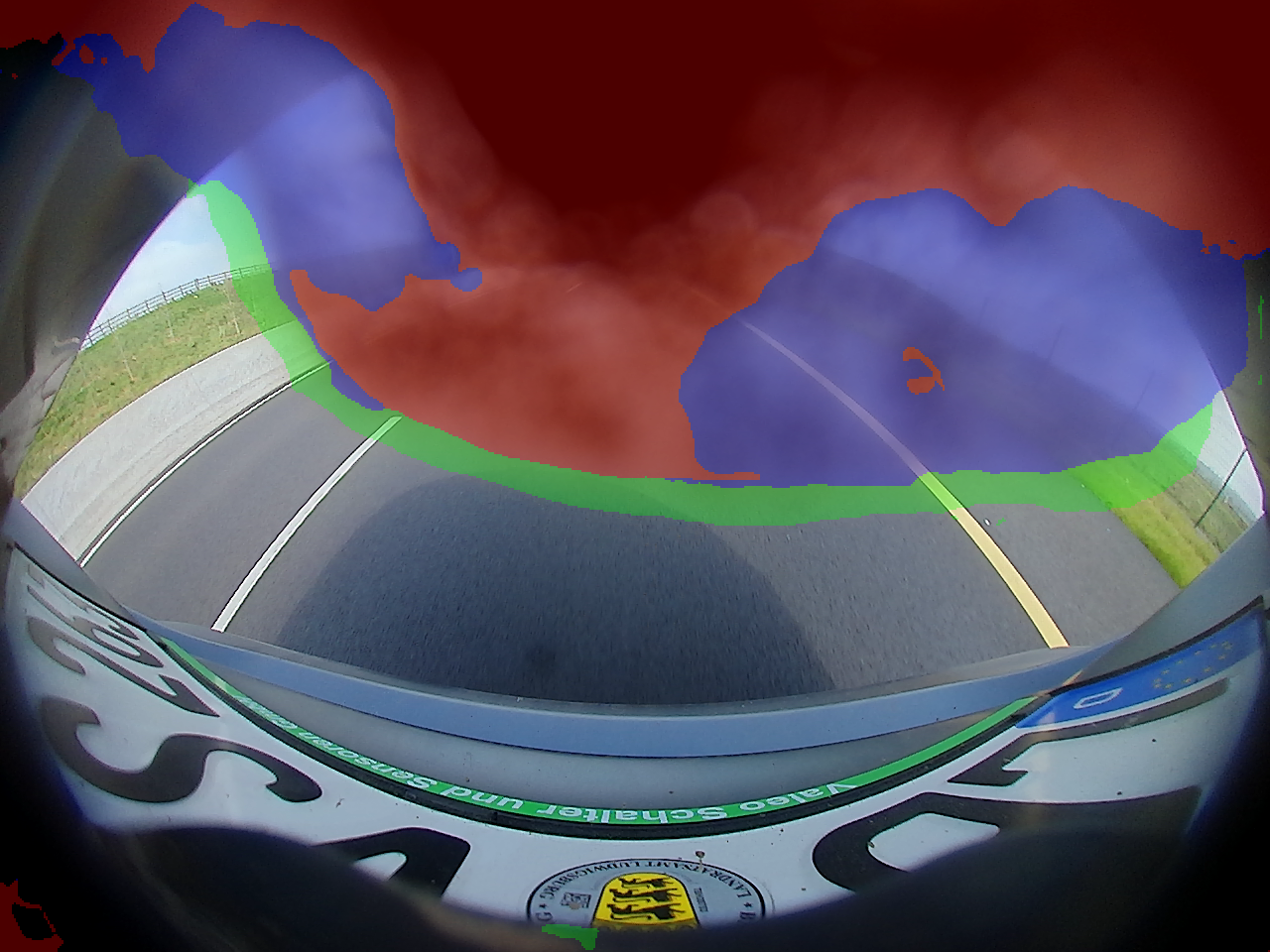}
\includegraphics[height=0.15\textheight]{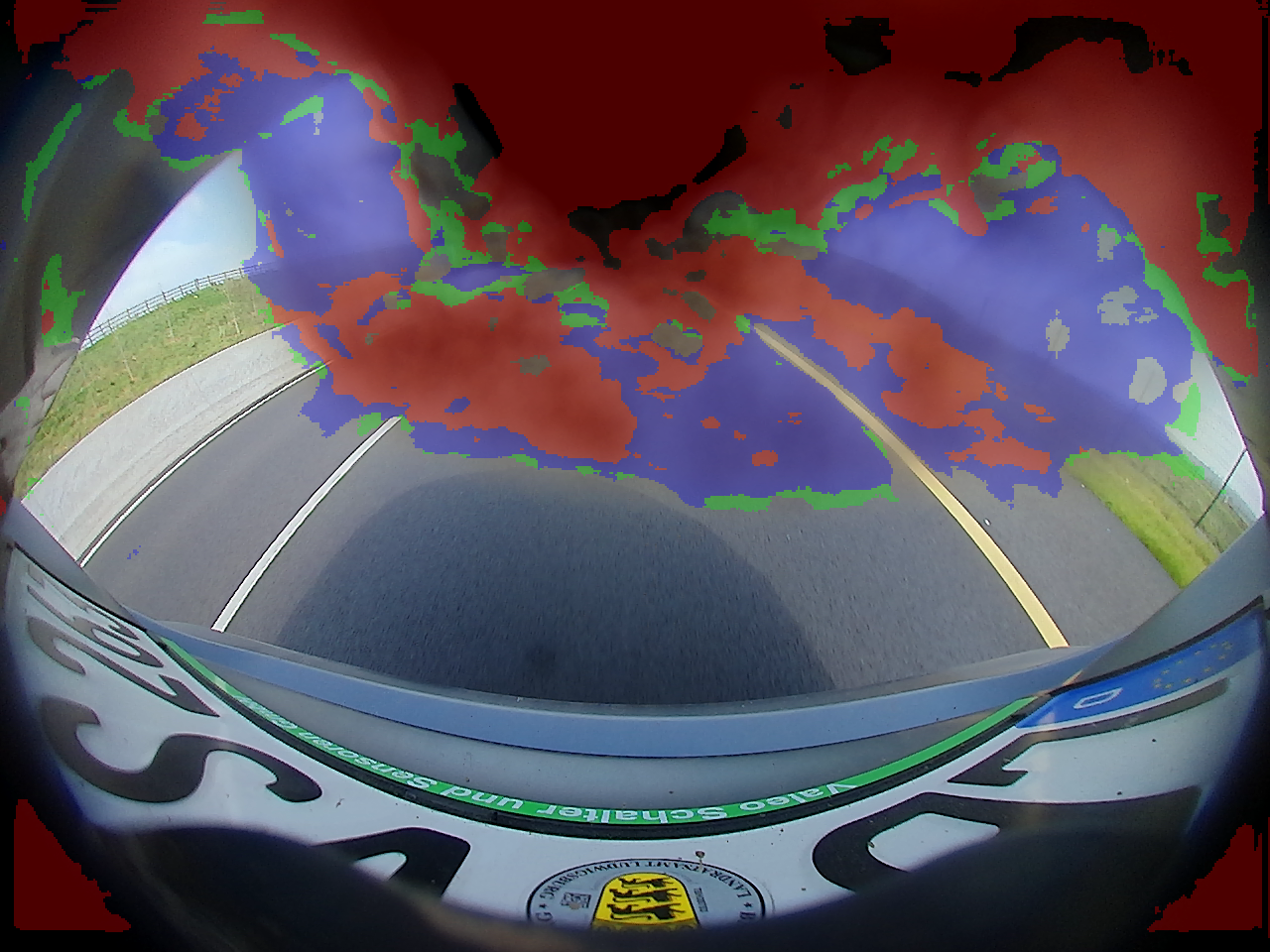}
\includegraphics[height=0.15\textheight]{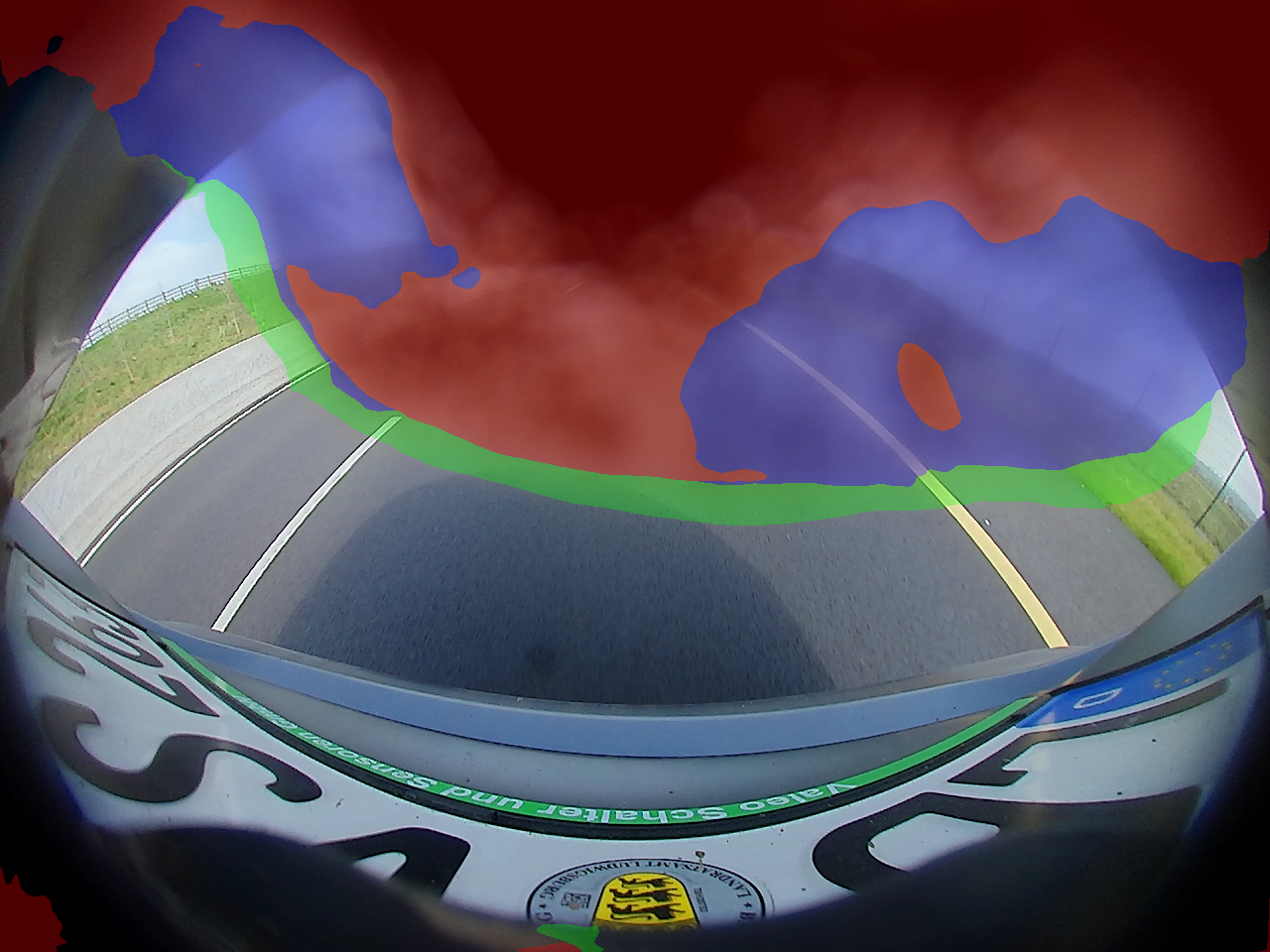}
\includegraphics[height=0.15\textheight]{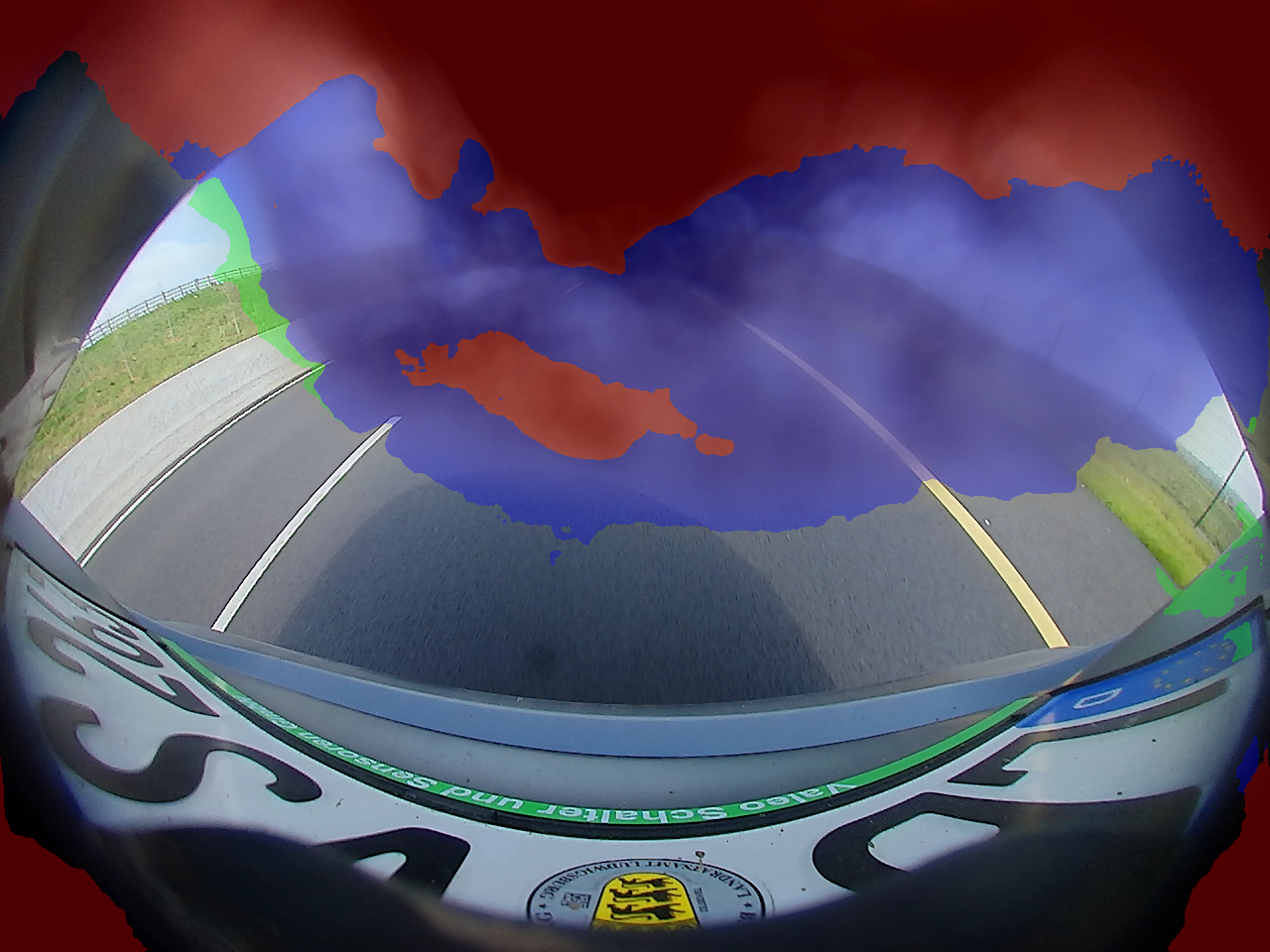}

\vspace{1.5mm}

\includegraphics[height=0.15\textheight]{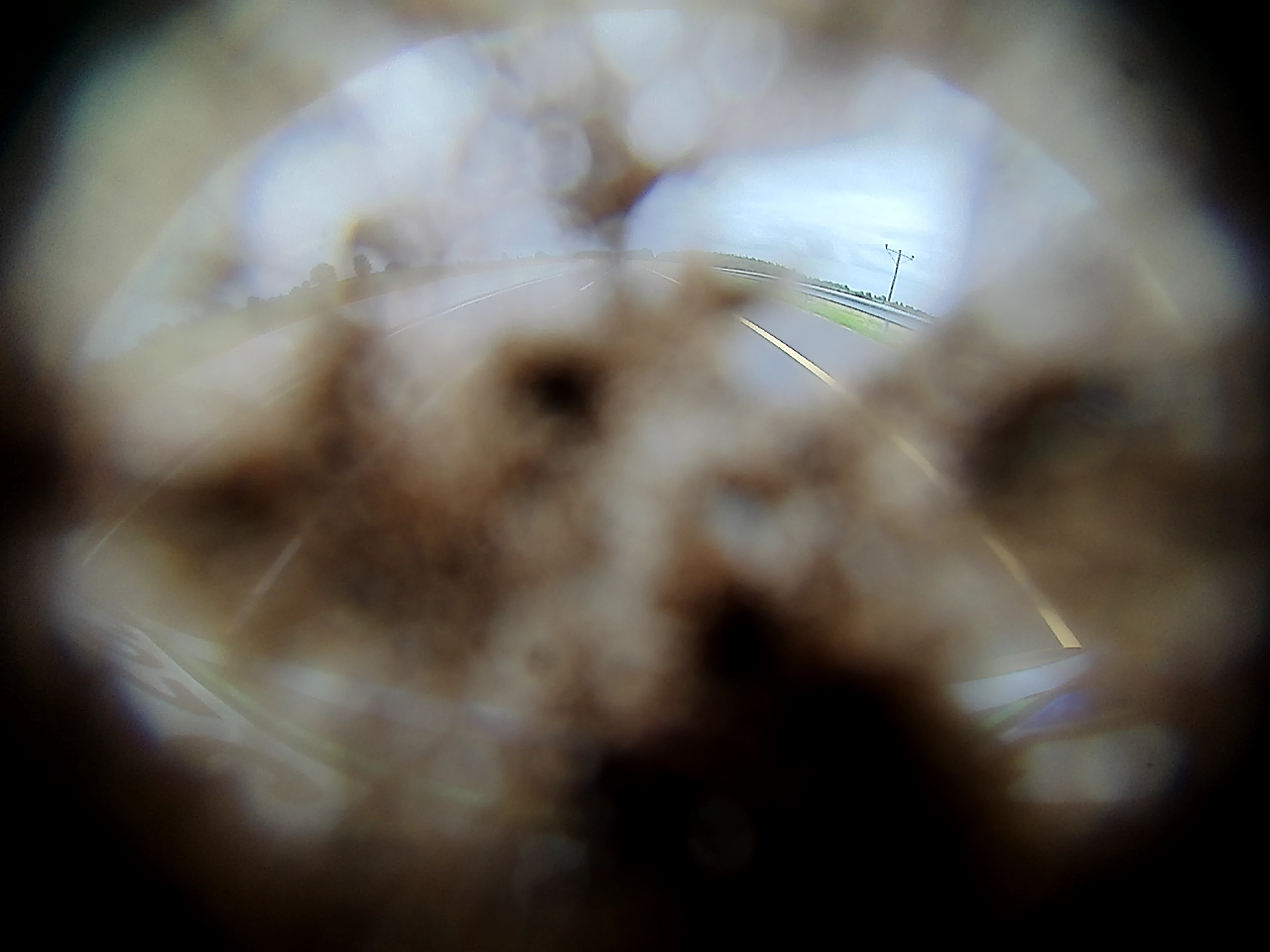}
\includegraphics[height=0.15\textheight]{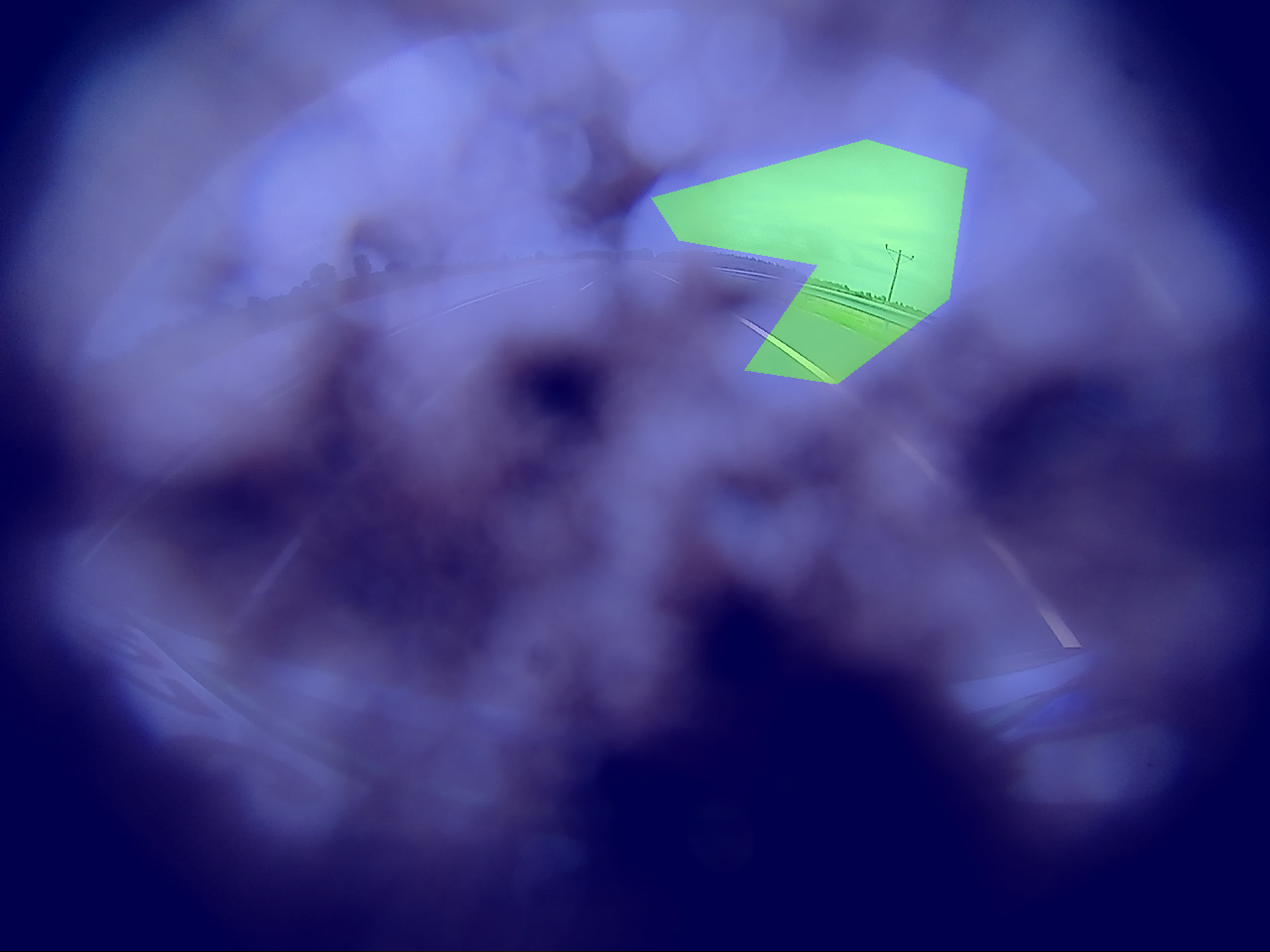}
\includegraphics[height=0.15\textheight]{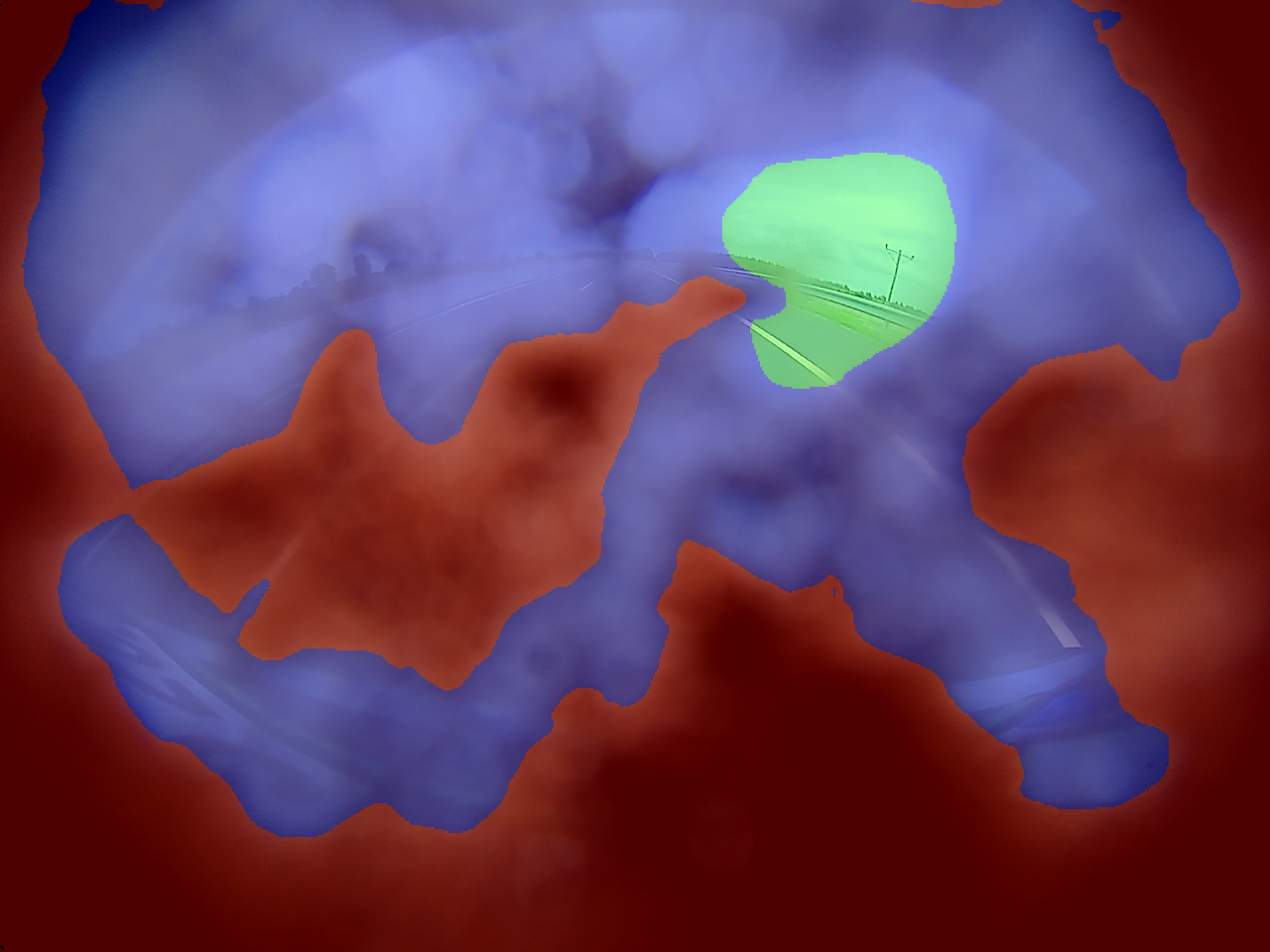}
\includegraphics[height=0.15\textheight]{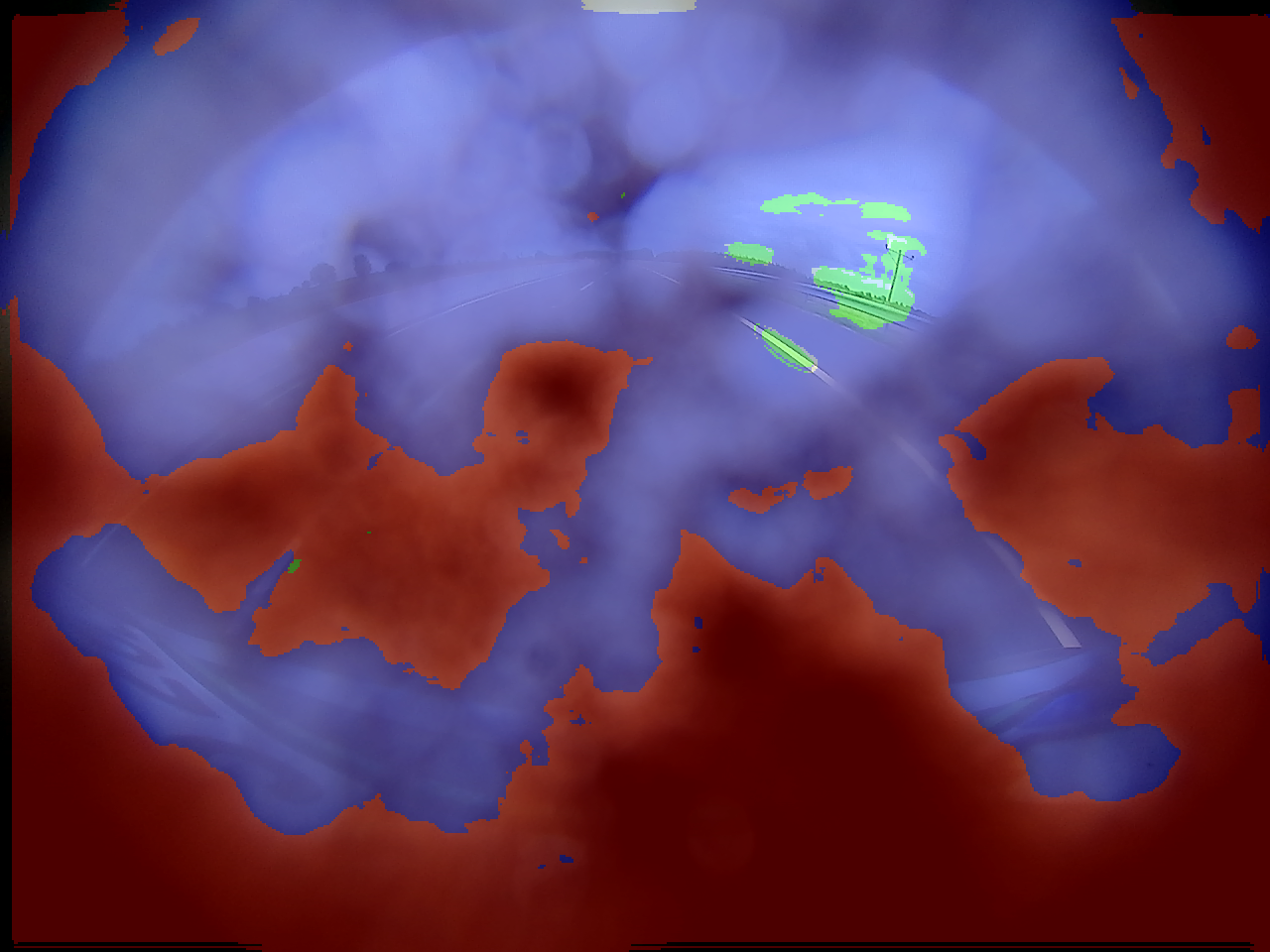}
\includegraphics[height=0.15\textheight]{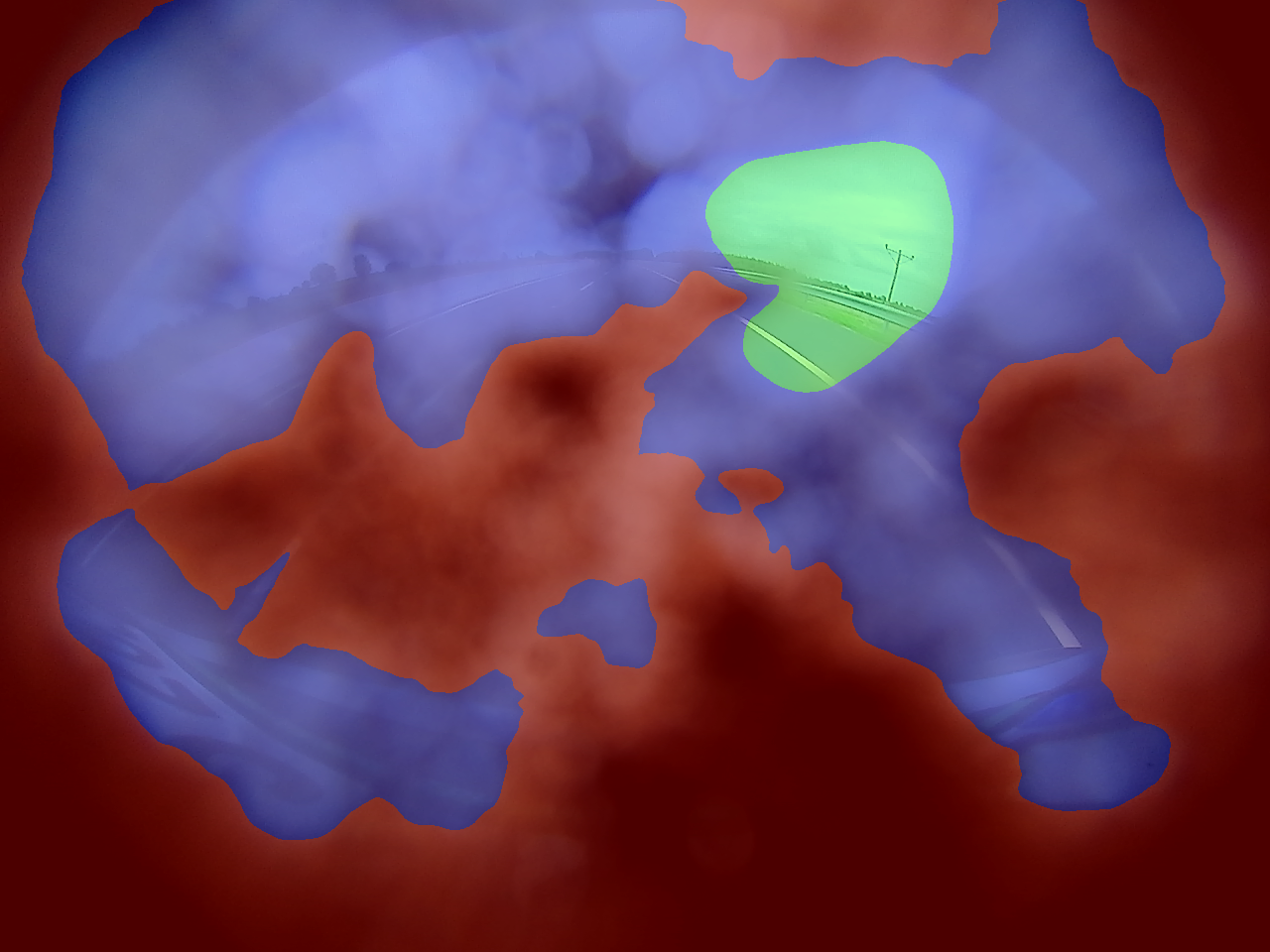}
\includegraphics[height=0.15\textheight]{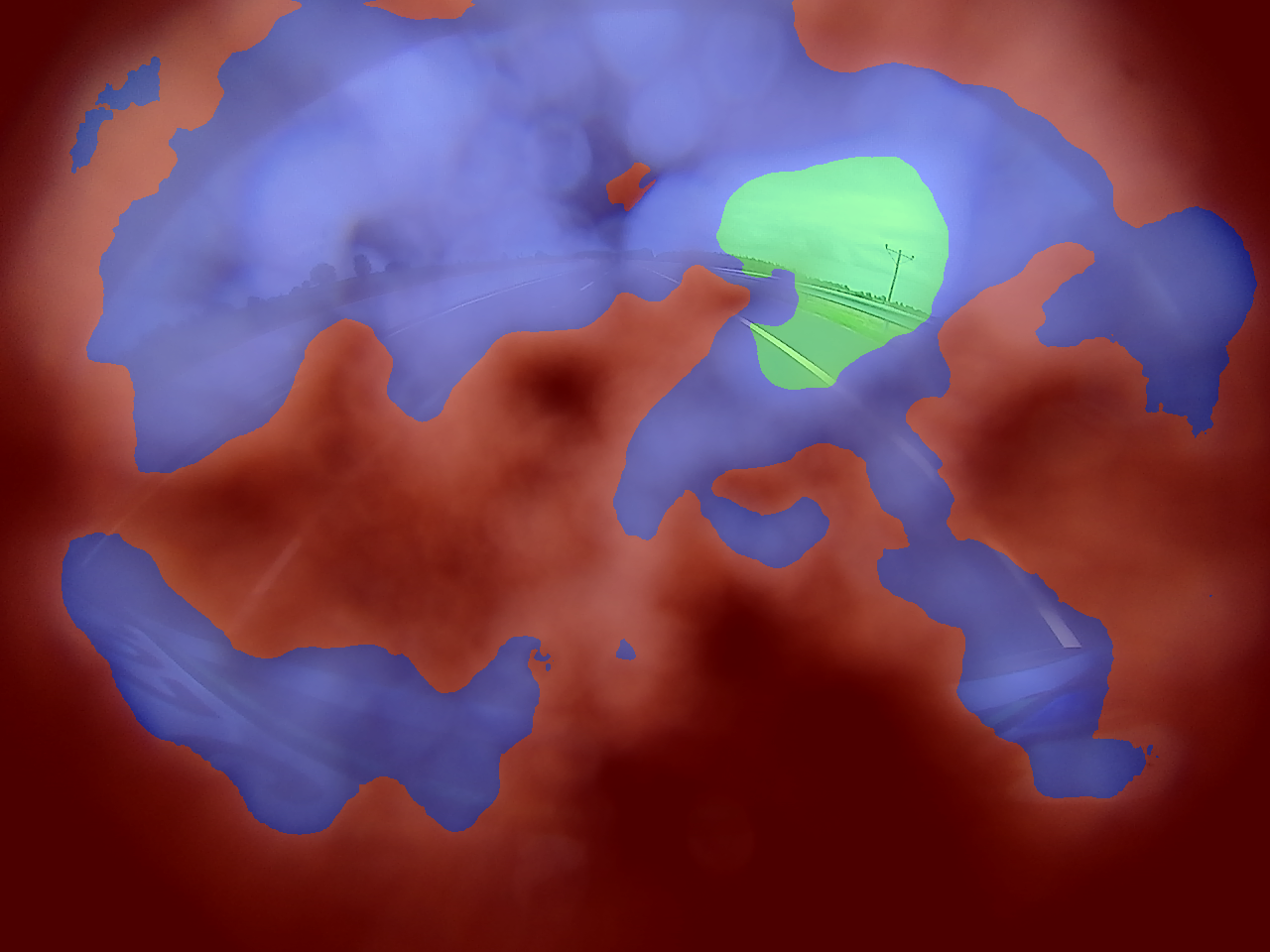}

\caption{Three sets of qualitative results. Top row:  Original image, manual annotation and ensemble refined annotation. Next row: Individual classifier predictions (ensemble input) from MaskSegNet, DeepLabV3+ with the ResNet and DeepLabV3+ with the Xception backbone. Only 4 out of 9 pseudo-label outputs are shown here due to limited space. Class color coding: {\color{green}Green --- transparent}, {\color{blue}blue --- semi-transparent}, {\color{red} red --- opaque}, and original image color --- clean. }
\label{fig:classifier-results}
\end{figure*}

We use $9$ PLs in total. First one is the original manual annotation, which is quite rough (polygonal annotation for soiling parts, cf. Figure~\ref{fig:soilingannotation}) and error-prone (cf. Figure~\ref{fig:classifier-results}). Next $3$ PLs are formed by classification via DeepLabV3+~\cite{deeplabv3plus2018} network with ResNet50~\cite{he2016deep} backbone with $3$ different approaches for calculating the inference, namely: multi-scale prediction, sliding window prediction and holistic prediction. Another $3$ PLs are built analogically by classification output of DeepLabV3+ but this time using the inception~\cite{inception} backbone. The last $2$ PLs consist of the output of an in-house simple soiling segmentation network called MaskSegNet, where the first one is using a model trained on the quarter of the original image resolution ($\mathcal{I}^{H/4 \times W/4}$) and the second is using a model trained on the half of the original image resolution ($\mathcal{I}^{H/2 \times W/2}$). All networks were trained using the non-public part of the soiling part of the WoodScape dataset~\cite{Yogamani_2019_ICCV}. In Figure~\ref{fig:classifier-results}, we show several examples of the $4$ out of the $9$ PLs, so the reader can compare their quality.

We describe the full learning scheme in Algorithm~\ref{alg:algo}. It consists of two consecutive stages. In the first stage, one of the $9$ pseudo-labels is randomly selected as the correct label and the ensemble network is trained to minimize the cross-entropy loss. The second stage uses the same architecture as the first stage, but this time the nearest neighbor pseudo-label to the prediction of the first stage classifier is selected as the correct label. The second stage network is also trained to minimize the cross-entropy loss. The ensemble network architecture is outlined in Figure~\ref{fig:EnsembleSAWSeg}.

Since the variance of the images and the soiling patterns is not very high, we use the common trick applicable to semantic segmentation networks by selecting a random crop of the image and corresponding crop of its annotation. This also allows us to use standard data augmentation, such as random rotation or horizontal/vertical flipping.


\section{Experimental Evaluation} \label{sec:experiments}

In this section, we summarize the conducted experimental evaluation and obtained results.

\subsection{Experimental Setup}

We made use of an internal dataset consisting of $20$k samples. They were split into train/val/test datasets in the ratio of $6:2:2$ based on stratified sampling techniques discussed in~\cite{uricar2019challenges}. To evaluate the refined (i.e. ensemble generated) annotations, two PSPnet~\cite{zhao2017pyramid} models with ResNet-50~\cite{he2016deep} backbone were trained. One model was trained with the manual annotations and the second one with the refined annotations. Both models are trained under the same configuration settings like optimizer, number of epochs, etc. 

\begin{figure}[tb]
    \centering
    \includegraphics[width=0.5\textwidth]{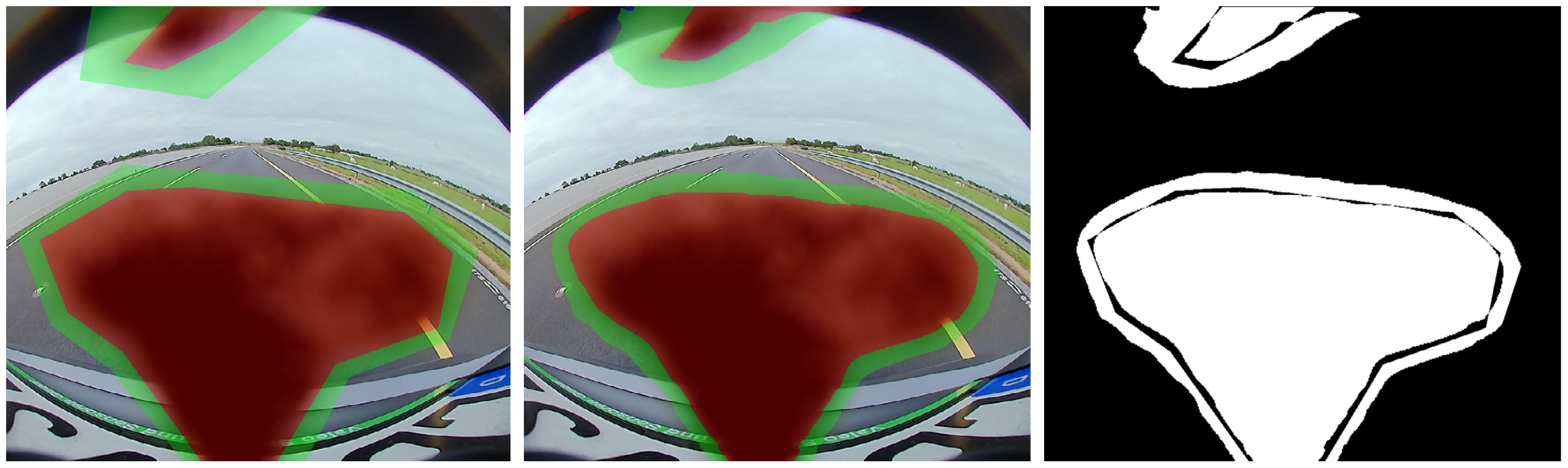}
    \caption{Left to Right: Manual annotation, ensemble refined annotation and the mask indicating the intersection area.}
    \label{fig:eval_annotation_sets}
\end{figure}

The networks are trained using the train/val sets and evaluated using the test set. Key Performance Indicators (KPIs) are computed for both the networks using three different annotation variants for the same test set. The first annotation variant composes of the original manual annotations, the second one composes of the refined, ensemble generated, annotations, and the third annotation set was generated by computing the intersection between the manual and the ensemble labels where only pixels annotated with the same class in both annotations were considered (Figure~\ref{fig:eval_annotation_sets}). This intersection area pixels indicate the model performance on regions where both the manual annotations and ensemble annotations are aligned. Two metrics are used for the comparison: i) IoU (Intersection over Union) between the ground truth (GT) and the prediction, ii) accuracy of the prediction.

\subsection{Quantitative results} 

Table~\ref{tab:qual-results} summarizes the obtained results in terms of IoU and prediction accuracy. The discussion of the results follows.

\paragraph{Model trained on manual annotation}
Mean IoU score for this model on manual annotations is $65.26\%$. While the model performs well on clean and opaque classes it has relatively low score for transparent and semi-transparent classes. This is understandable as the transparent and semi-transparent classes are hard to annotate due to fuzziness in the definition, resulting in noisy annotations.

The mean IoU obtained with the ensemble annotation test set ($74.89 \%$) is considerably higher than the one obtained when testing on the manual annotations ($65.26 \%$). For both annotations the model performance remains quite similar and improvements are observed in noisy categories of classes. This shows that there is a label noise in transparent and semi-transparent refined annotations compared to manual ones. This conclusion can hold further better,
\begin{itemize}
\item If both the models perform better on ensemble annotations, showing that ensemble generated annotations are less noisy 
\item If the model trained on ensemble annotations outperforms the model trained on manual annotations while evaluated on manual annotations and ensemble annotations, indicating relatively better generalization.  
\end{itemize}

From the Table \ref{tab:qual-results} the overall IoU of the manual annotations trained model is better on the ensemble test set, it is important to highlight the increase in performance per classes: transparent ($+16.21 \%$), semi-transparent ($+15.26 \%$) and opaque ($+6 \%$). The same conclusion can be drawn regarding the mean accuracy computed ($+9.1 \%$).

\begin{table*}[tb]
\caption{Qualitative results of the two models trained with manual or ensemble annotations. Each model has been tested on 3 different test sets (Manual, Ensemble  or Intersection test sets).  Accuracies obtained on Ensemble test set are higher than when testing on manual test set for both networks. The Intersection test sets allows us to compare the performance of the 2 networks on a common domain and shows that the network trained on Ensemble annotations performs better.}
\label{tab:qual-results}
\centering
\begin{tabular}{|l|l|l|l|l|l|l||l|l|l|l|l|l|}
\hline
\textbf{Model} & \multicolumn{6}{c||}{Trained on Manual Annotations} & \multicolumn{6}{c|}{Trained on Ensemble Annotations} \\ \hline
\textbf{Annotation} & \multicolumn{2}{c|}{Manual} & \multicolumn{2}{c|}{Ensemble} & \multicolumn{2}{c||}{Intersection} & \multicolumn{2}{c|}{Manual} & \multicolumn{2}{c|}{Ensemble} & \multicolumn{2}{c|}{Intersection} \\ \hline
\textbf{Class} & IoU $\uparrow$ & Acc $\uparrow$ & IoU & Acc & IoU & Acc & IoU & Acc & IoU & Acc & IoU & Acc \\ \hline
clean & $94.36$ & $96.72$ & $95.41$ & $96.37$ & $86.15$ & $86.44$ & $94.11$ & $96.82$ & $\mathbf{96.18}$ & $\mathbf{97}$ & $86.69$ & $86.93$ \\ \hline
transparent & $35.77$ & $47.55$ & $51.98$ & $68.57$ & $38.21$ & $73.89$ & $\mathbf{37.92}$ & $51.17$ & $\mathbf{60.71}$ & $78.37$ & $42.76$ & $\mathbf{83.53}$ \\ \hline
semi\_transparent & $49.52$ & $68.01$ & $64.78$ & $82.23$ & $50.7$ & $87.41$ & $\mathbf{51.26}$ & $69.2$ & $\mathbf{74.13}$ & $88.54$ & $56.09$ & $\mathbf{92.74}$ \\ \hline
opaque & $81.39$ & $93.71$ & $87.39$ & $95.22$ & $81.97$ & $\mathbf{98.44}$ & $81.15$ & $91.87$ & $\mathbf{90.25}$ & $95.14$ & $84.06$ & $97.93$ \\ \hline
Mean & $65.26$ & $76.5$ & $74.89$ & $85.6$ & $64.26$ & $86.55$ & $\mathbf{66.11}$ & $77.26$ & $\mathbf{80.32}$ & $89.76$ & $67.4$ & $\mathbf{90.28}$ \\ \hline
\end{tabular}
\end{table*}

\begin{table}[tb]
\centering
\caption{Manual visual inspection score on test set.}
\label{tab:v_inspection-results}
\begin{tabular}{|l|l|l|l|}
\hline
 & Manual & Ensemble & Both \\ \hline
Reviewer 1 & $10.39$ & $46.59$ & $43.01$ \\ \hline
Reviewer 2 & $16.79$ & $39.64$ & $43.57$ \\ \hline
Reviewer 3 & $16.61$ & $41.88$ & $41.52$ \\ \hline
\textbf{Average} & $14.59$ & $42.70$ & $42.70$ \\ \hline
\end{tabular}
\end{table}

\paragraph{Model trained on ensemble annotation}
The model trained on ensemble annotations improved by $+5.43 \%$ in mean IoU score, compared to the manual annotations trained model when testing on ensemble test set. This model performs slightly better than the model trained on manual annotations overall (mean IoU score). This shows that the model trained on ensemble annotations can generalize well on manual annotations as well. More importantly, this model outperforms the manual annotation model by $+2.15 \%$ and $+1.72\%$ for transparent and semi-transparent classes respectively.

\paragraph{Comparison of results computed on Intersection set}
So far we have shown results on inter and intra annotation domains for both models. But it is also important to compare the performance of the models on a common domain. To compare both the networks on a common domain, we evaluated both models on an ``intersection'' test set (see Figure \ref{fig:eval_annotation_sets}). Intersection test set labels are generated as the pixels that shares the same class in both manual and ensemble generated annotations, respectively. 
The obtained results show that the network trained on the ensemble annotation performs better than the other one. While the IoU for the class ``clean'' stays the same, the IoU for the three other classes increased significantly for the model trained on the ensemble generated annotations (transparent: $+4.55 \%$, semi-transparent: $+5.39 \%$ and opaque: $2.09 \%$). Along with the IoU KPI, the overall accuracy increased as well ($+3.73 \%$).
Hence, our conclusion that the ensemble generated annotations contain less label noise compared to the manual annotations is proven via the evaluation of the model performance on segmentation tasks.


\subsection{Manual visual inspection}

Since the goal of the proposed ensemble classifier is fixing the annotations, we cannot rely solely on the numbers presented in the previous paragraph. Therefore, we conducted an experiment with a manual inspection of the generated annotations. Manual inspection was done as an AB test by a team of $3$ image annotation experts with the knowledge of the soiling detection task. Tests are carried using a Python based software tool, where the original image along with its manual annotation and its generated ensemble annotation were showed (see Figure~\ref{fig:manual-vs-ensemble}). Each reviewer visually evaluated $280$ images and had to select which annotation is better in his opinion. Table~\ref{tab:v_inspection-results} shows the score obtained after this manual inspection. The percentage of the better manual annotation is summarized in the first column, the better ensemble annotation is shown in the second column, and of similar annotations in the third column. It can be observed that in $42.70 \%$ of the cases, ensemble annotations where considered better than the manual annotation ($14.59 \%$). It is important to notice, that in the $42.7 \%$ of the annotation that where found similar, clean images have been reviewed too. The number of images manually inspected covers $20 \%$ of the hold out test set.



\section{Conclusions} \label{sec:conclusions}

Camera lens soiling annotations are difficult to annotate precisely due to less defined boundaries and semantic ambiguities. In this work, we propose to refine the manual coarse polygonal annotations to fine precise annotations using an ensemble pseudo-label model. We demonstrate significantly improved annotations using manual visual verification and quantitative results based on a soiling segmentation model. It also illustrates that coarse manual annotations are sufficient to significantly lowering the annotation costs. 
Precise annotations are crucial for robust models and we demonstrate a semi-supervised approach to improve the annotations. In future work, we plan to jointly train the ensemble annotation refiner and the learning model in an active learning loop.


{\small
\bibliographystyle{IEEEtran}
\bibliography{bib/references}
}

\end{document}